\def\year{2021}\relax
\documentclass[letterpaper]{article} 
\usepackage{aaai21}  
\usepackage{times}  
\usepackage{helvet} 
\usepackage{courier}  
\usepackage[hyphens]{url}  
\usepackage{graphicx} 
\usepackage{booktabs}
\usepackage{soul}
\usepackage{multirow}
\usepackage{pervasives}
\usepackage{amsmath}
\usepackage[switch]{lineno}  %

\usepackage{natbib}  
\usepackage{caption} 
\title{Exploring Transfer Learning For End-to-End Spoken Language Understanding}
\author {

        Subendhu Rongali\textsuperscript{\rm 1, 2}\thanks{Work done during the author's summer internship.}
        Beiye Liu \textsuperscript{\rm 1}
        Liwei Cai \textsuperscript{\rm 1}
        Konstantine Arkoudas \textsuperscript{\rm 1} \\
        Chengwei Su \textsuperscript{\rm 1}
        Wael Hamza \textsuperscript{\rm 1} \\
}
\affiliations {
    \textsuperscript{\rm 1} Amazon Alexa AI, New York
    \textsuperscript{\rm 2} University of Massachusetts Amherst  \\
    srongali@cs.umass.edu, \{beiyeliu, cliwei, arkoudk, chengwes, waelhamz\}@amazon.com
}

\begin{document}

\maketitle

\begin{abstract}
Voice Assistants such as Alexa, Siri, and Google Assistant typically use a two-stage Spoken Language Understanding pipeline; first, an Automatic Speech Recognition (ASR) component to process customer speech and generate text transcriptions, followed by a Natural Language Understanding (NLU) component to map transcriptions to an actionable hypothesis. An end-to-end (E2E) system that goes directly from speech to a hypothesis is a more attractive option. These systems were shown to be smaller, faster, and better optimized. However, they require massive amounts of end-to-end training data and in addition, don't take advantage of the already available ASR and NLU training data.

In this work, we propose an E2E system that is designed to jointly train on multiple speech-to-text tasks, such as ASR (speech-transcription) and SLU (speech-hypothesis), and text-to-text tasks, such as NLU (text-hypothesis). We call this the Audio-Text All-Task (AT-AT) Model and we show that it beats the performance of E2E models trained on individual tasks, especially ones trained on limited data. We show this result on an internal music dataset and two public datasets, FluentSpeech and SNIPS Audio, where we achieve state-of-the-art results. Since our model can process both speech and text input sequences and learn to predict a target sequence, it also allows us to do zero-shot E2E SLU by training on only text-hypothesis data (without any speech) from a new domain. We evaluate this ability of our model on the Facebook TOP dataset and set a new benchmark for zeroshot E2E performance. We will soon release the audio data collected for the TOP dataset for future research.

\end{abstract}

\begin{figure*}[t]
\centering
\hspace*{2.5em}\includegraphics[width=1.8\columnwidth]{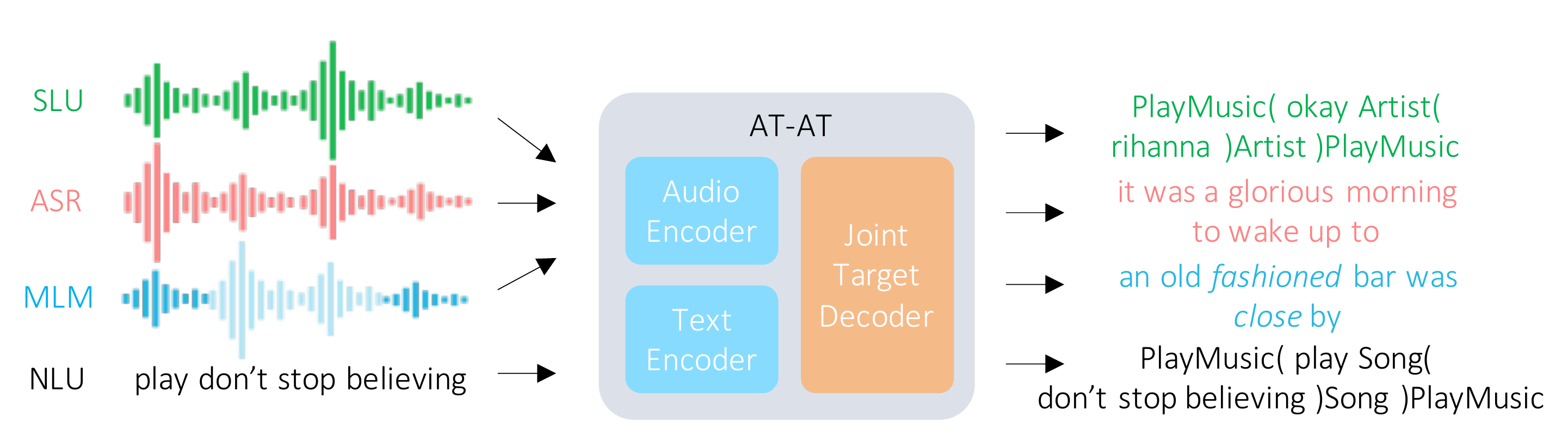}
\caption{Pretraining AT-AT with audio-to-text and text-to-text tasks. The audio and text inputs go to separate encoders but share a joint decoder, which decodes the target sequence based on the task. Task labels are passed as BOS tokens while decoding.}
\label{at-at-w}
\end{figure*}

\section{Introduction}
In recent years, there has been a dramatic surge in the adoption of voice assistants such as Amazon Alexa, Apple Siri, and Google Assistant. Customers use them for a variety of tasks such as playing music and online shopping.

These voice assistants are built on complex Spoken Language Understanding (SLU) systems that are typically too large to store on an edge device such as a mobile phone or a smart speaker. Hence, user traffic is routed through a cloud server to process requests. This has led to privacy concerns and fueled the push for tiny AI and edge processing, where the user requests are processed on the device itself. 

Traditional SLU systems consist of a two-stage pipeline, an Automatic Speech Recognition (ASR) component that processes customer speech and generates a text transcription \emph{(ex. play the song watermelon sugar)}, followed by a Natural Language Understanding (NLU) component that maps the transcription to an actionable hypothesis consisting of intents and slots \emph{(ex. Intent: PlaySong, Slots: SongName - watermelon sugar)}. An end-to-end (E2E) system that goes directly from speech to the hypothesis would help make the SLU system smaller and faster, allowing it to be stored on an edge device. It could potentially also be better optimized than a pipeline since it eliminates cascading errors.

However, E2E systems are not used in practice because they have some key issues. These systems are hard to build since they consist of large neural components such as transformers and require massive amounts of E2E training data. They also don't make use of the vastly available training data for the ASR and NLU components that could be used to enhance their performance, because the examples in these datasets may not be aligned to create an E2E training sample. Another issue is feature expansion, a scenario where a new domain, with new intents and slots, is added to the voice assistant's capabilities. Here, developers typically only have access to some synthetically generated text-hypothesis examples. Speech data isn't readily available and it is very expensive to collect. E2E models thus fail as they require lots of new audio and hypothesis data to learn this new domain.

In this work, we build an E2E model that mitigates these issues using transfer learning. We call it the Audio-Text All-Task (AT-AT) Model. AT-AT is an E2E transformer-based model that is jointly trained on multiple audio-to-text and text-to-text tasks. Examples of these tasks include speech recognition (ASR), hypothesis prediction from speech (SLU), masked LM prediction (MLM), and hypothesis prediction from text (NLU). Our model achieves this by converting data from all these tasks into a single audio-to-text or text-to-text format. Figure~\ref{at-at-w} shows this joint training phase in detail. Our findings indicate that there is significant knowledge transfer taking place from multiple tasks, which in turn helps in downstream model performance. We see that the AT-AT pretrained model shows improved performance on SLU hypothesis prediction on internal data collected from Alexa traffic. We also report state-of-the-art results on two public datasets: FluentSpeech \cite{Lugosch2019SpeechMP}, and SNIPS Audio \cite{Saade2018SpokenLU}. 

Furthermore, since our model contains a text encoder, it can consume both audio and text inputs to generate a target sequence. By jointly training on both audio-to-text and text-to-text tasks, we hypothesize that this model learns a shared representation for audio and text inputs. This allows us to simply train on new text-to-text data and get audio-to-text performance for free, giving us a way to do E2E hypothesis prediction in a zero-shot fashion during feature expansion. We test this approach on an internal dataset from Alexa traffic, and an external dataset, Facebook TOP \cite{gupta2018semantic}. Since TOP consists of only text data, we collected speech data for the test split using an internal tool at Amazon. We will soon release this dataset.

In summary, our contributions are as follows.
\begin{itemize}
    \item We developed an E2E SLU model that is jointly trained on multiple audio-to-text and text-to-text tasks and shows knowledge transfer and SLU performance improvements.
    \item We report state-of-the-art results on two public SLU datasets, FluentSpeech and SNIPS Audio.
    \item We show how to perform zero-shot E2E hypothesis prediction with our model.
    \item We report a new benchmark for zeroshot E2E SLU on the Facebook TOP dataset and will soon release the test data.
\end{itemize}

\section{Related Work}
The architecture of prior E2E SLU models is taken from neural speech recognition literature. Speech recognition was originally performed using hidden Markov models that predict acoustic features, followed by word-level language models \cite{Furui2000DigitalSP}. More recently, deep learning models have become more popular for this task \cite{Hinton2012DeepNN}. Deep learning models solve this task by posing it as a sequence-to-sequence problem \cite{Graves2013SpeechRW,Nassif2019SpeechRU}. With the success of transformer-based sequence-to-sequence models on text based tasks \cite{vaswani2017attention}, researchers have explored and shown success in applying them for speech recognition \cite{Mohamed2019TransformersWC,Karita2019ACS}. Our architecture is based on these models.

Other end-to-end SLU models also closely resemble this sequence-to-sequence encoder-decoder framework \cite{Haghani2018FromAT,Lugosch2019SpeechMP}. The slot-filling task for SLU is formulated as a target text sequence by wrapping the target English tokens with intent and slot tags, which was shown to achieve state of the art results \cite{Rongali_2020}. Our approach improves upon these models by introducing transfer learning. The transfer learning paradigm we adopt here is similar to prior efforts that use multiple tasks or pretraining to improve SLU performance \cite{wang2020large,jia2020large}. The audio-text shared training idea also has prior work. However, these efforts require parallel audio-text data \cite{denisov2020pretrained}, or are evaluated on a simpler classification task \cite{sari2020training}.

Zeroshot E2E SLU, where we only have text NLU training data but no audio has also been explored. Recently, \cite{Lugosch2020UsingSS} approached this task using speech synthesis. They generate synthetic speech from text using a Text to Speech (TTS) system and use the resultant audio to train their models. While this approach is simple and intuitive, its success greatly depends on access to a good TTS system. We propose a method that can perform this task, end-to-end, without any TTS system, and can also be used in conjunction with a TTS system to further improve performance.

Finally, an important part of all these models is the representation of audio. The raw audio waveform is typically converted into higher level features before being passed to the actual models. While Mel-Frequency Cepstral Coefficitents (MFCC) have been the traditional choice for this conversion, Log-filterbank features (LFB) have become more popular recently \cite{fayek2016}. We use LFB features here.

\section{The AT-AT Model}
In this section, we explain the design of our proposed Audio-Text All-Task (AT-AT) model. AT-AT is trained to jointly perform multiple audio-to-text and text-to-text tasks. 

We hypothesize that AT-AT will benefit from potential knowledge transfer in a multi-task setting. This is in line with findings in a recent work \cite{Raffel2019ExploringTL} that converts a variety of text based natural language tasks into source and target text sequences and shows knowledge transfer by using a single shared sequence-to-sequence model. AT-AT can also be used as a pretrained checkpoint to build end-to-end models on new datasets to achieve better performance. Finally, we believe that AT-AT is a powerful audio-text shared representation model that would allow us to do E2E zeroshot prediction using just text data. 

When training AT-AT with audio tasks, the input audio signal is pre-processed to obtain a sequence of LFB features, which is taken as the source sequence. For text tasks, the source sequence is simply the text input tokens. The target consists of a sequence of tokens corresponding to the task being solved. For example, the target sequence is a sequence of words if the task is speech recognition. If the task is SLU or NLU hypothesis prediction, the target consists of the intent and slot tags as well as the words within them, a formulation based on recent work that solves this task as a sequence-to-sequence problem \cite{Rongali_2020}. An example set of source-target sequences for tasks is shown in Figure~\ref{at-at-w}. We pass the task label as the beginning-of-sequence (BOS) token in the target decoder. This way, the model can conditionally decode the target sequence based on the observed input and the task being solved. Note that previous multi-task text-to-text models \cite{Raffel2019ExploringTL} add this information to the source sequence itself. Since our source sequence can be in the audio space, we add the task label at the start of the target sequence. 

While the audio encoder trained on multiple audio-to-text tasks presents an obvious transfer learning advantage in SLU, our reasons for incorporating a text encoder in this model are two-fold; first, we can add more text-to-text tasks in the pretraining phase, and second, more importantly, this would enable us to train on a task with only text-to-text data and expect good audio-to-text performance. AT-AT thus has the ability to do zero-shot end-to-end SLU by training on only annotated text data, an important ability that comes in handy during feature expansion, where new intents and slots need to be added to the model without any audio data available. This situation arises because the text data for new intents and slots can be synthetically generated but the audio data is not readily available and is expensive to collect.

A model can develop the zero-shot ability if the audio and text inputs share a common space from which the target sequence is generated. A common way to learn a shared space from two input sources is to explicitly impose an L2 loss penalty on the hidden state vectors of the two aligned input sources \cite{denisov2020pretrained}. This is however infeasible in our setup because the hidden states from the audio and text input sequences are not single vectors, but sequences of vectors of different lengths and resolution. While we can pool these vectors to get a single vector, doing so would result in a huge information bottleneck which makes the decoder incapable of decoding the target sequence well. We resolve this problem by avoiding the explicit vector alignment altogether, hence eliminating any need to pool the encoder hidden states. We use a single shared decoder to process the hidden state vectors of both the audio and text encoder. By constraining the complexity of this decoder, we force it to learn a shared representation between audio and text so that it can solve both tasks without solving them separately.

AT-AT consists of two phases of training: 1) the pretraining phase, where we train our model on multiple audio-to-text and text-to-text tasks, and 2) the finetuning phase, where we finetune our model on a single downstream task. The architecture of AT-AT, these two phases, and our zeroshot end-to-end approach are described below.

\subsubsection{Architecture}

\begin{figure}[t]
\centering
\includegraphics[width=\columnwidth]{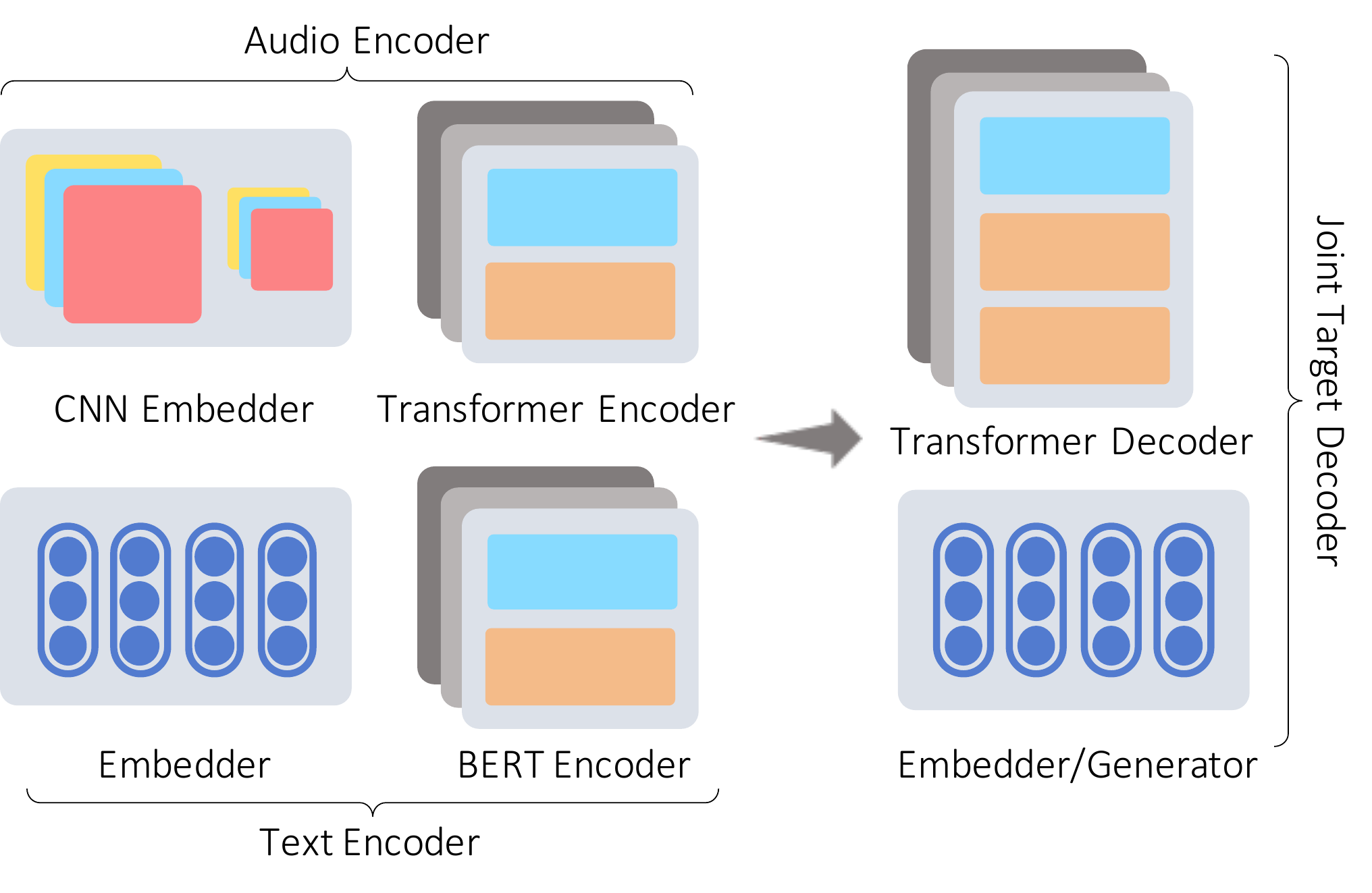}
\caption{Model Components of AT-AT}
\label{at-at-mc}
\end{figure}

AT-AT has an architecture similar to many transformer-based speech recognition models proposed recently \cite{Karita2019ACS,Mohamed2019TransformersWC}, which contain an encoder-decoder framework to process a source audio sequence and decode the target text sequence. In addition to the audio encoder, our model also contains a text encoder to process text sequences.

The audio encoder consists of multiple convolutional and max pooling layers to contextually embed the audio LFB spectrogram. This is followed by a transformer encoder \cite{vaswani2017attention}. These convert the input audio sequence into a much shorter sequence of hidden states to be consumed by the decoder. For the text encoder, we use BERT \cite{devlin2018bert}, which consists of an embedder to embed the input tokens and their positions, followed by a transformer encoder. The text encoder typically produces hidden states that are larger in size than the audio encoder so we use a projection layer to project the text hidden states down to match the dimensionality of the audio hidden states. Once this is done, both the text and audio sequences generate a sequence of hidden states of the same size. 

We use a single transformer decoder to decode the targets from the sequence of encoder hidden states. Both the text and audio inputs go through the same generation process, which allows the model to learn a shared representation without any explicit loss penalty to align them.

We use byte-pair encoding (BPE) to split the target words into smaller pieces. We only split the target English words, not any tokens corresponding to intent and slot tags. The target sequence tokens are embedded using a standard embedding matrix. The transformer decoder consumes the current token embedding and performs a multi-head multi-layer attention over the encoder hidden states to generate a decoder hidden state. The decoder hidden state is passed through a generator layer that shares weights with the embedding matrix. The generator layer assigns a probability mass to each token in the target vocabulary, representing the probability of that token being generated next. Further details on this decoder framework are beyond the scope of this paper and can be found in \cite{vaswani2017attention}. Note that instead of a fixed BOS token to start decoding as usual, we use the task label as the BOS token. Figure~\ref{at-at-mc} lays out these components.

\subsubsection{Pretraining Phase}

The pretraining phase of AT-AT consists of training with multiple audio-to-text and text-to-text sequence-to-sequence tasks. Examples from all tasks are randomly sampled in each batch during pretraining. Figure~\ref{at-at-w} shows the pretraining phase in action, where we train with three audio-to-text tasks: SLU hypothesis prediction (SLU), automatic speech recognition (ASR), masked-audio LM prediction (MLM), and one text-to-text task: NLU hypothesis prediction (NLU). For the MLM task, the audio corresponding to certain words is masked out and the model is trained to predict the whole target sequence. We perform audio-word alignment prior to the masking using an external tool; more details on this are in the datasets section. We require at least one audio-to-text and one text-to-text task if the model will be used to do zeroshot E2E prediction.

\subsubsection{Finetuning Phase}

In the finetuning phase, we start from the pretrained model and train it on a specific downstream task, such as SLU. We hypothesize that pretraining with multiple tasks allows the model to transfer knowledge from different tasks, allowing it to be better regularized and obtain a warm start for optimization for the downstream task.

When the pretrained model is used as a starting point for new datasets with new intents and slots, unseen target token embeddings are randomly initialized. The model is first trained by freezing all pretrained parameters so that these new parameters get to a good optimization point. They are then gradually unfrozen over time as the model is finetuned.

\subsubsection{Zeroshot End-to-End}

In the zeroshot scenario, we have access to a new annotated text-to-text dataset and we want to construct an E2E model capable of predicting the target sequence given audio input. It is a common occurrence in the feature expansion phase in voice assistants, where a new domain is added to the voice assistant's capabilities. For example, say a voice assistant is currently capable of handling user requests in music and shopping domains. We want to add the capability for it to handle requests in a new domain, say books, such as reading a book. In this case, developers usually write down some launch phrases and annotate them to perform a certain task in the new domain. The audio data for these phrases doesn't exist yet. The goal is to bootstrap an E2E model that can process audio data and generate the hypothesis by just training on the text data.

AT-AT allows us to do this easily when it is pretrained on a certain task from both audio and text inputs. In the voice assistant feature expansion case for example, the pretraining phase is carried out with an SLU task on existing domains, an NLU task on existing domains, and any other tasks we want to add such as ASR and MLM. Once the pretraining is complete, we simply finetune the model using the annotated text NLU data from the new domain and test on audio data.

While this is one way to train E2E models without audio data, another way is to simply generate the missing audio data using a Text-to-Speech (TTS) system and use it for training. This approach is however contingent on the availability of a good TTS system. With AT-AT, we can perform zeroshot prediction without a TTS system. Moreover, when we do have access to a TTS system, we can add the generated synthetic audio to the finetuning phase and finetune AT-AT on both the synthetic audio and text. We hypothesize that this is better than simple E2E training since the text NLU data helps train the language model within the decoder even better, allowing AT-AT to work harmoniously with synthetic audio to further improve performance.  

\section{Evaluation}
\subsection{Datasets}
Our experiments are carried out on a combination of internal and publicly available datasets. We describe them here.

\subsubsection{Internal SLU Data}
Our internal dataset is created by sampling utterances from user traffic of our voice assistant, Alexa. This is done in compliance with user commitments with regards to privacy and anonymity.  We select only utterances from the music domain for the first set of experiments. This dataset contains about 3M training utterances, 100k validation, and 100k testing utterances comprising 23 intents and 95 slots. Each utterance here contains the audio, text transcript, and the SLU hypothesis.

For our low-resource experiments, we sample 10\% of utterances from the above dataset and select the audio and hypotheses. We pick the text transcriptions from the rest to create data for the ASR task during AT-AT pretraining. 

\subsubsection{LibriSpeech ASR Data}
We also compile an ASR dataset by downloading all splits of the publicly available LibriSpeech dataset \cite{Panayotov2015LibrispeechAA}, giving us $\sim$1000 hours of data. This data is comprised of multiple speakers reading sentences from audio books in the LibriVox project.

\subsubsection{MLM Data}
We create the dataset for the MLM task by modifying the LibriSpeech dataset. We first use an external audio alignment tool, Gentle\footnote{https://github.com/lowerquality/gentle} that is built on the Kaldi framework \cite{Povey2011TheKS}. Once this is done, we mask 15\% of the words in each transcript and the corresponding audio in the audio file. This masked audio is then processed to produce the LFB features to produce the audio input and the target sequence is the entire transcript.

\subsubsection{Public SLU Datasets}
We also evaluate AT-AT on public SLU datasets to compare with the state-of-the-art results. We use two public datasets: FluentSpeech \cite{Lugosch2019SpeechMP}, and SNIPS Audio \cite{Saade2018SpokenLU} in our evaluation. The FluentSpeech dataset consists of target sequences that are 3-tuples, not sequences. We convert them into target sequences using some pre-processing rules to create data in the required format. 
There are about 23k train, 3k valid, and 4k test examples in this dataset.

The annotations in SNIPS are in the form of intents and slots, so can be trivially converted into target sequences in the required format. We use the smart-lights close-field and far-field datasets from the SNIPS dataset for our experiments and report results with 5-fold cross validation since there are no explicitly delineated train-test splits. These dataset are extremely small, each consisting of a total of 1660 examples.

\subsubsection{Zeroshot SLU Datasets}
For our zeroshot experiments, we require text NLU training data and audio SLU test data on an unseen domain. We collect two datasets for this. First is an internal dataset that consists of utterances sampled from Alexa traffic in the books domain. We extract around 200k text NLU training examples, and 10k audio SLU test examples comprising 21 intents and 47 slots. 

We also construct a zeroshot dataset from the publicly available Facebook TOP \cite{gupta2018semantic} dataset. This is a challenging dataset that contains complex utterances with nested intents and slots. It contains $\sim$32k train, 4k eval, and 9k test utterances. We want to evaluate the performance of AT-AT on this dataset to show its effectiveness in a complete domain shift. With TOP, we use the training and validation data splits as is. Using an internal utterance collection tool, we collected audio data for a fraction of the test split, about 1915 utterances from multiple speakers, to test zeroshot performance. We will soon release this dataset \footnote{In Progress. Awaiting legal approval at Amazon.}. 

For the zeroshot experiments, one of our baselines is an E2E model built by generating synthetic speech data from a TTS System. We use Amazon Polly\footnote{https://aws.amazon.com/polly/} as our TTS system. We use 9 randomly selected speakers and the neural engine to create speech data for utterances.

\begin{table}
\begin{tabular}{lcc}
\toprule
\bf Method & \bf{SemER} & \bf{EM Accuracy} \\
\midrule
E2E Model w. 100\% data & \emph{baseline} & \emph{baseline} \\
E2E Model w. 10\% data & +8.63 & -11.82 \\
\midrule
AT-AT, Pretrained (10\%) & +2.13 & -2.52 \\
AT-AT, Finetuned (10\%) & \bf +1.45 & \bf -1.49 \\
\bottomrule
\end{tabular}
\caption{
    Results on the low-resource music dataset. 
}
\label{tab:music1}
\end{table}

\begin{table}
\begin{tabular}{lcc}
\toprule
\bf Method & \bf{SemER} & \bf{EM Accuracy} \\
\midrule
E2E Model w. 100\% data & \emph{baseline} & \emph{baseline} \\
Prod ASR, linear chain CRF & +0.61 & -1.02\\
Prod ASR, BiLSTM + CRF & -0.45 & +0.70 \\
\midrule
AT-AT, SLU + ASR & \bf -1.16 & \bf +1.39 \\
AT-AT, SLU + ASR + MLM & -1.01 & +1.23 \\
\bottomrule
\end{tabular}
\caption{
    Results on the full music dataset. 
}
\label{tab:music2}
\end{table}

\subsection{Experimental Details}
We use 80-dim LFB features to process the audio signals. The target English words were tokenized using byte-pair encoding to obtain a final vocabulary of 5k.

We use a 2-layer 2D CNN with 256 final units and a transformer encoder with 12 layers, 4 heads, 256 units, and 2048 hidden units as our audio encoder. The text encoder is the standard BERT-base encoder \cite{devlin2018bert}. The target decoder consists of a 256-dim tied embedding/generator matrix and a transformer decoder with 6 layers, 4 heads, 256 units, and 2048 hidden units. We use noam learning rate schedule with 4000 warm-up steps and an adam optimizer with learning rate 1. We use cross entropy loss with label smoothing ($\epsilon=0.1$) as our loss function. During inference, we use beam search with a beam-size of 4.

When finetuning with gradual unfreezing, we use a learning rate multiplier of 0 for the first 500 steps, and 0.2, 0.5, 0.7 for the next 100 steps each, finally reaching 1 after 800 steps and training normally from there on. We didn't perform extensive hyper-parameter tuning for our experiments. 

\subsection{AT-AT in Low Resource Settings}
Our first set of experiments evaluate the effect of AT-AT multi-task training on improving the performance of an E2E model trained on a low-resource annotated dataset. To simulate the low resource setting, we take our internal music SLU dataset and sample 10\% of the data to obtain the speech and SLU annotations. For the rest of the examples, we obtain the ASR transcripts to create the ASR dataset for the multi-task training. Our AT-AT model is pretrained on these two tasks. We evaluate this model's performance on the test set immediately after pretraining. We then perform the finetuning step on just the 10\% SLU data and perform another evaluation.
\subsubsection{Baselines}
We train two E2E models as baselines. These models have the same architecture as our AT-AT model, without the multi-task component or the text encoder. The first model is trained on the full internal music SLU dataset. The second model is trained on the extracted 10\% dataset. We expect our AT-AT model, that makes use of the additional ASR data from music to recuperate any drop in performance between these two models. 
\subsubsection{Results}
Table \ref{tab:music1} shows the results of these experiments. We report two metrics here, the semantic error rate (SemER), and the exact match (EM) accuracy. Exact match accuracy simply corresponds to the accuracy obtained by matching the entire predicted hypothesis to the gold hypothesis. SemER is a more slot-filling oriented metric that rewards partially correct hypotheses. It is an internal metric that is used to evaluate the performance of SLU models built for Alexa. Given the number of insertion (I), deletion (D), and substitution (S) errors in the predicted hypothesis, it is given by $\frac{\text{S} + \text{I} + \text{D}}{\text{\# total slots + 1 (for intent)}}$. 
We want a lower SemER and a higher EM accuracy. Due to internal regulations, we do not report the absolute numbers on internal datasets. For this experiment, we use the performance numbers of the E2E model trained on 100\% data as the baseline and report the remaining numbers relative to it. 

We observe that there is a big drop in performance when we train a model on 100\% data vs 10\% data. The SemER increases by 8.63 absolute points. However, our AT-AT model, pretrained with additional ASR data recuperates most of this performance, mitigating this increase in error to only 2.23 points. Finetuning on the SLU data further improves performance, giving us a error increase of just 1.45 points. We see a similar trend in the exact match accuracy scores as well where our models lose the least number of accuracy points. These results show that multi-task training with additional ASR data is hugely beneficial in a low-resource scenario.

\subsection{Building Better E2E Models with AT-AT}
The previous experiment showed that the performance of an E2E model trained on a low-resource dataset (10\% data) can be improved by adding additional ASR data and training in a multi-task setting with AT-AT. In this experiment, we want to take this a step further and evaluate if we can improve the performance of a model trained on the full 100\% dataset using any available external data. We pretrain AT-AT with the full 100\% SLU dataset and in addition, include two more tasks: ASR and MLM. We use the LibriSpeech ASR and MLM datasets as described in the datasets section for these two tasks. Note that these datasets are from a completely different domain than music. We want to determine whether we can improve the performance of an E2E model by adding tasks from other domains with transferable knowledge. 

We evaluate our model in two settings. The first setting consists of pretraining with 2 tasks, SLU and ASR, followed by finetuning on the 100\% SLU dataset. The second setting's pretraining phase consists of 3 tasks, SLU, ASR, and MLM, followed by finetuning again on the 100\% SLU dataset. Our baseline is the E2E model trained on 100\% music SLU data from the previous set of experiments. For context, we also report numbers from two 2-stage pipeline models for SLU. We use a production-level ASR system from Amazon for the first stage. For the second (NLU) stage, we experiment with a linear chain CRF and a pretrained BiLSTM + CRF (SOTA). The BiLSTM + CRF model beats transformer-based models for this dataset \cite{Rongali_2020}.

\subsubsection{Results}
We report the results of these experiments in Table \ref{tab:music2}. We again report relative numbers here since this is in an internal dataset. We use the performance of the E2E 100\% model as the baseline. We see that adding LibriSpeech ASR data and pretraining with AT-AT improves SemER on the internal music SLU test set by 1.16 points, representing a significant relative error reduction. The exact match accuracy also improves by 1.4 absolute points here. With all three tasks, we see that the SemER improves by 1.01 points, slightly worse than the previous number. We believe the lack of further improvement from the MLM task might be because it doesn't contribute new information to the model.

\begin{table}
\begin{tabular}{lccc}
\toprule
\multirow{2}{*}{\bf{Method}} & \multirow{2}{*}{\bf{HypER}} & \multicolumn{2}{c}{\bf{EM Accuracy}} \\
& & \bf{Hyp} & \bf{Full} \\ 
\midrule
E2E Model & 8.3 & 91.7 & 83.4 \\
SOTA 1 \cite{Lugosch2019SpeechMP} & 1.2 & 98.8 & -- \\
SOTA 2 \cite{wang2020large} & 1.0 & 99.0 & -- \\
AT-AT & \bf 0.5 & \bf 99.5 & \bf 99.0 \\
\bottomrule
\end{tabular}
\caption{
    Results on the FluentSpeech dataset. 
}
\label{tab:flsp}
\end{table}

\begin{table}
\begin{tabular}{lccc}
\toprule
\multirow{2}{*}{\bf{Method}} & \multicolumn{2}{c}{\bf{EM Accuracy}} \\
& \bf{Hyp} & \bf{Full} \\ 
\midrule
\multicolumn{3}{c}{Close Field}\\
\midrule
SNIPS \cite{Saade2018SpokenLU} & 84.22 & -- \\
Google \cite{Saade2018SpokenLU} & 79.27 & -- \\
E2E Model & \multicolumn{2}{c}{\emph{no convergence}} \\
E2E Model w. pret. AT-AT encoder & 81.87 & 53.90 \\
AT-AT & \bf 84.88 & \bf 66.51 \\
\midrule
\multicolumn{3}{c}{Far Field}\\
\midrule
SNIPS \cite{Saade2018SpokenLU} & 71.67 & -- \\
Google \cite{Saade2018SpokenLU} & 73.43 & -- \\
E2E Model & \multicolumn{2}{c}{\emph{no convergence}} \\
E2E Model w. pret. AT-AT encoder & 67.83 & 38.92 \\
AT-AT & \bf 74.64 & \bf 53.25 \\
\bottomrule
\end{tabular}
\caption{
    Results on the SNIPS dataset.
}
\label{tab:snips}
\end{table}

\begin{table*}
\centering
\begin{tabular}{lccccc}
\toprule
\bf Method & \bf{EM Accuracy} & \bf Precision & \bf Recall & \bf{F1} & \bf{Tree validity}\\
\midrule
E2E Model w. Syn. Audio &  69.19 & 67.24 & 65.15 & 66.18 & 98.85\\
AT-AT zeroshot & 51.54 & 51.31 & 49.80 & 50.55 & 98.96 \\
AT-AT zeroshot + Syn. Audio & \bf 70.60 & \bf 67.98 & \bf 66.39 & \bf 67.18 & \bf 99.37\\
\bottomrule
\end{tabular}
\caption{
    Results on the TOP dataset. 
}
\label{tab:top}
\end{table*}

\subsection{AT-AT on Public Datasets}
In this set of experiments, we evaluate how AT-AT's pretraining can help improve performance on other datasets. We selected the publicly available FluentSpeech and SNIPS Audio datasets to compare to state-of-the-art models.

We use the AT-AT model pretrained with 2 tasks from the previous experiment and finetune it on the FluentSpeech and SNIPS datasets. We also trained end-to-end models from scratch on these two datasets. To perform an ablation on the AT-AT finetuning approach, we report an additional number on the SNIPS dataset, for a model that uses a pretrained AT-AT audio encoder. This model, compared to the full AT-AT model would give us an idea of how much the decoder pretraining helps, in addition to the encoder pretraining.

\subsubsection{Baselines}
For the FluentSpeech dataset, we compare to two SOTA models. The first model is the best model from \cite{Lugosch2019SpeechMP}. It is a multi-layer RNN-based network, with lower layers trained to predict aligned word targets from the LibriSpeech dataset. The final task is formulated as a 3-way classification task, not a generation task like our AT-AT model. The second model is a transformer-based pretrained model from \cite{wang2020large}.

For the SNIPS Audio dataset, we compare with the two models reported in \cite{Saade2018SpokenLU}, SNIPS and Google. The SNIPS model consists of a pipe-lined approach with an acoustic model for ASR, followed by a language model, and slot tagging model for NLU. The Google model is from Google's DialogFlow cloud service\footnote{https://cloud.google.com/dialogflow}.

\subsubsection{Results}
Table \ref{tab:flsp} reports the results on the FluentSpeech dataset. We report error rate on the complete hypothesis (HypER), exact match accuracy on the hypothesis, and the exact match accuracy on the full target sequence. While the end-to-end model doesn't perform too well from scratch, our AT-AT finetuned model beats the state-of-the-art model by 0.5 accuracy points. This corresponds to a 50\% error reduction.

Table \ref{tab:snips} contains the results on the SNIPS dataset. We report the exact match accuracy on the hypothesis and the full target sequence here. We see that our AT-AT pretrained models have the best performance on both the close-field and far-field sets with a 5-fold cross validation setup. They beat both Google and SNIPS models' numbers previously reported. We also see that the AT-AT model is vastly superior to an end-to-end model with a pretrained audio encoder. This is especially evident with the accuracy scores on the full target sequence where the AT-AT model beats it by 10-15 absolute points. Note that we weren't able to train an end-to-end model from scratch due to extremely small dataset size.

\begin{table}
\begin{tabular}{lcc}
\toprule
\bf Method & \bf{SemER} & \bf{EM Accuracy} \\
\midrule
E2E Model w. Real Audio & \emph{baseline} & \emph{baseline} \\
\midrule
E2E Model w. Syn. Audio & +5.05 & -9.58 \\
AT-AT zeroshot & +11.90 & -15.14 \\
AT-AT zeroshot + Syn. Audio & \bf +3.31 & \bf -5.20 \\
\bottomrule
\end{tabular}
\caption{
    Results on the books dataset. 
}
\label{tab:books}
\end{table}

\subsection{Zeroshot E2E with AT-AT}
In the final experiments, we evaluate the performance of AT-AT on zeroshot end-to-end tasks. Here, we only have text training data and we want to evaluate on speech.

We first pretrained AT-AT on 4 tasks: SLU (speech-to-hypothesis), ASR, MLM, and NLU (text-to-hypothesis). We use data from the internal music dataset (for SLU and NLU), and the LibriSpeech dataset for ASR and MLM. This model is then finetuned on the internal books dataset and the Facebook TOP dataset with the text NLU training data as described in the architecture section. We also finetune AT-AT in another setting, using text NLU training data along with the synthetic speech data from our TTS system. We want to show that in addition to performing zeroshot prediction without access to a TTS system, we can also work with an existing TTS system to further improve performance. 

\subsubsection{Baselines}
For the internal books dataset, we built an E2E model on real audio training data to obtain a rough upper bound and gauge the zeroshot performance. In addition to this, we also trained another E2E model on the synthetic dataset constructed from our TTS system. 

For the TOP dataset, we don't have real audio training data, so our baseline was an E2E model trained on the synthetic training data. For a fair comparison to AT-AT, we use the same pretrained audio encoder for these E2E models.

\subsubsection{Results}
Tables \ref{tab:top} and \ref{tab:books} report results of the zeroshot experiments. On the internal books dataset, we report relative numbers for SemER and EM accuracy. We use the performance numbers of an E2E model trained on real speech data as baselines. Using synthetic training data gives us a SemER of \emph{baseline} + 5.05 points . AT-AT achieves a zeroshot SemER of \emph{baseline} + 11.90 without access to a TTS system, a respectable number compared to the aforementioned model. AT-AT when finetuned with additional synthetic speech data beats an E2E model trained on only synthetic data, obtaining a SemER of \emph{baseline} + 3.31 (lowest increase in error).

On the TOP dataset, we report all the recommended metrics given in the dataset but we are primarily interested in exact match accuracy. Note that our test set was compiled by recording speech for a fraction of the full test set. We observe the same trend here that we observe on the books dataset. An E2E model trained on synthetic data achieves an accuracy of 69.19 while AT-AT achieves 51.54. While there is a significant drop, it is to be noted that AT-AT sees absolutely no new labeled audio instances, giving it a significant disadvantage while switching input models during inference. It also doesn't require a TTS system for this training. The E2E model is however beaten by the AT-AT model finetuned with additional synthetic data, which achieves an accuracy of 70.60 (2\% relative improvement).

\section{Conclusion}
We propose the Audio-Text All-Task (AT-AT) model that uses transfer learning to improve the performance on end-to-end SLU. AT-AT beat the performance of E2E models on our internal music data, both in the full and low-resource settings. It also achieved state-of-the-art performance on the FluentSpeech (99.5\% EM Accuracy) and SNIPS audio datasets (84.88\% close-field, 74.64\% far-field EM) with significant improvements over prior models. AT-AT also demonstrated its ability to perform zeroshot E2E SLU, without access to a TTS system, and by learning a shared audio-text representation without any explicit loss penalty to force the audio and text hidden states into the same space. We also showed how AT-AT can work in conjunction with a TTS system to further improve E2E performance. It achieves a zeroshot E2E EM Accuracy of 70.60 on the TOP dataset. 

On a closing note, we would like to remark that AT-AT somewhat mimics actual human learning. We typically read a lot more words than we hear. But when we hear a word for the first time, we transfer our knowledge of that word from when we read it. AT-AT similarly learns to understand and perform NLU tagging from text and then applies this knowledge when it is given speech.

\bibliography{aaai.bib}

\end{document}